\documentclass[a4paper]{article}

\usepackage{INTERSPEECH2018}

\usepackage{tabularx}
\usepackage{enumitem}

\newcolumntype{Y}{>{\centering\arraybackslash}X}
\newcolumntype{Z}{>{\centering\arraybackslash}X}

\title{An Efficient Approach to Encoding Context for Spoken Language
Understanding}
\name{Raghav Gupta, Abhinav Rastogi, Dilek Hakkani-T{\"u}r}
\address{Google AI, Mountain View}
\email{raghavgupta@google.com, abhirast@google.com, dilek@ieee.org}

\begin{document}

\maketitle

\begin{abstract}
  In task-oriented dialogue systems, spoken language understanding, or SLU,
  refers to the task of parsing natural language user utterances into semantic
  frames. Making use of context from prior dialogue history holds the key to
  more effective SLU. State of the art approaches to SLU use memory networks to
  encode context by processing multiple utterances from the dialogue at each
  turn, resulting in significant trade-offs between accuracy and computational
  efficiency. On the other hand, downstream components like the dialogue state
  tracker (DST) already keep track of the dialogue state, which can serve as a
  summary of the dialogue history. In this work, we propose an efficient
  approach to encoding context from prior utterances for SLU. More specifically,
  our architecture includes a separate recurrent neural network (RNN) based
  encoding module that accumulates dialogue context to guide the frame parsing
  sub-tasks and can be shared between SLU and DST. In our experiments, we
  demonstrate the effectiveness of our approach on dialogues from two domains.
\end{abstract}


\section{Introduction}
Task-oriented dialogue systems assist users with accomplishing tasks, such as
making restaurant reservations or booking flights, by interacting with them in
natural language. The capability to identify task-specific semantics is a key
requirement for these systems. This is accomplished in the spoken language
understanding (SLU) module, which typically parses natural language user
utterances into semantic frames, composed of user intent, dialogue acts and
slots ~\cite{tur2011spoken}, that can be processed by downstream dialogue system
components. An example semantic frame is shown for a restaurant reservation
query in Figure~\ref{fig:clu-example}. It is common to model intent, dialogue
act and slot prediction jointly ~\cite{hakkani2016multi, zhang2016joint,
liu2016attention, xu2013convolutional}, which is a direction we follow.

Much prior research into SLU has focused on single-turn language understanding,
where the system receives only the user utterance and, possibly, external
contextual features such as knowledge base annotations \cite{dhingra2017towards}
and semantic context from the frame \cite{dauphin2013zero}, as inputs. However,
task-oriented dialogue commonly involves the user and the system indulging in
multiple turns of back-and-forth conversation in order to achieve the user goal.
Multi-turn SLU presents different challenges, since the user and the system may
refer to entities introduced in prior dialogue turns, introducing ambiguity.
For example, depending on the context, ``three" could indicate a date, time,
number of tickets or restaurant rating. Context from previous user and system
utterances in multi-turn dialogue has been shown to help resolve these
ambiguities ~\cite{bhargava2013easy,xu2014contextual}. While initial work in
this direction used only the previous system turn for context, the advent of
deep learning techniques, memory networks \cite{sukhbaatar2015end} in
particular, facilitated incorporating context from the full dialogue history.

In essence, memory network-based approaches to multi-turn SLU store prior user
and system utterances and, at the current turn, encode these into embeddings,
using RNNs or otherwise. These memory embeddings are then aggregated to obtain
the context vector which is used to condition the SLU output at the current
turn. This aggregation step could use an attention mechanism based on cosine
similarity with the user utterance embedding ~\cite{chen2016end2end}. Other
approaches account for temporal order of utterances in the memory by using an
RNN for aggregation ~\cite{bapna2017sequential} or decaying attention weights
with time ~\cite{su2017how}.

Although improving accuracy, using memory networks for encoding context is not
computationally efficient for two reasons. First, at each turn, they process
multiple history utterances to obtain the SLU output. Secondly, dialogue context
could potentially be gleaned from existing dialogue system components such as
the dialogue state tracker ~\cite{henderson2014second, henderson2014word,
mrkvsic2017neural}. Using a separate SLU-specific network instead of reusing the
context from DST duplicates computation. Furthermore, such approaches work with
the natural language representation of the system utterance to have a consistent
representation with user turns, while ignoring the system dialogue acts, which
contain the same information but are more structured and have a smaller
vocabulary.

In this work, we investigate some effective approaches to encoding dialogue
context for SLU. Our contributions are two-fold. First, we propose a novel
approach to encoding system dialogue acts for SLU, substituting the use of
system utterances, which allows reuse of the dialogue policy manager's output to
obtain context. Second, we propose an efficient mechanism for encoding dialogue
context using hierarchical recurrent neural networks which processes a single
utterance at a time, yielding computational gains without compromising
performance. Our representation of dialogue context is similar to those used in
dialogue state tracking models ~\cite{yoshino2016dialogue, liu2017end,
williams2014web}, thus enabling the sharing of context representation between
SLU and DST.

\begin{figure}[t]
  \small
  \hspace{-7pt}
  \begin{tabular}{l c c c c c c}
    \textbf{System:} & \multicolumn{6}{l}{Which restaurant and for how many?}\\
    Dialogue Acts: & \multicolumn{6}{l}{\textit{request}(\#),
    \textit{request}(rest)}\\
    \textbf{User:} & Table & for & two & at & Olive & Garden\\
    & $\downarrow$ & $\downarrow$ & $\downarrow$ & $\downarrow$ & $\downarrow$ &
    $\downarrow$ \\
    Slot: & O & O & B-\# & O & B-rest & I-rest\\
    Intent: & \multicolumn{6}{l}{reserve\_restaurant}\\
    Dialogue Acts: & \multicolumn{6}{l}{\textit{inform}(\#),
    \textit{inform}(rest)}\\
  \end{tabular}
  \caption{An example semantic frame with slot, intent and dialogue act
  annotations, following the $IOB$ tagging scheme.}
  \label{fig:clu-example}
\end{figure}

The rest of this paper is organized as follows: Section \ref{sec:model}
describes the overall architecture of the model. This section also formally
describes the different tasks in SLU and outlines their implementation. Section
\ref{sec:experiments} presents the setup for training and evaluation of our
model. We conclude with experimental results and discussion.

\section{Approach}
\label{sec:model}
\begin{figure*}[t]
  \centering
  \includegraphics[width=\linewidth]{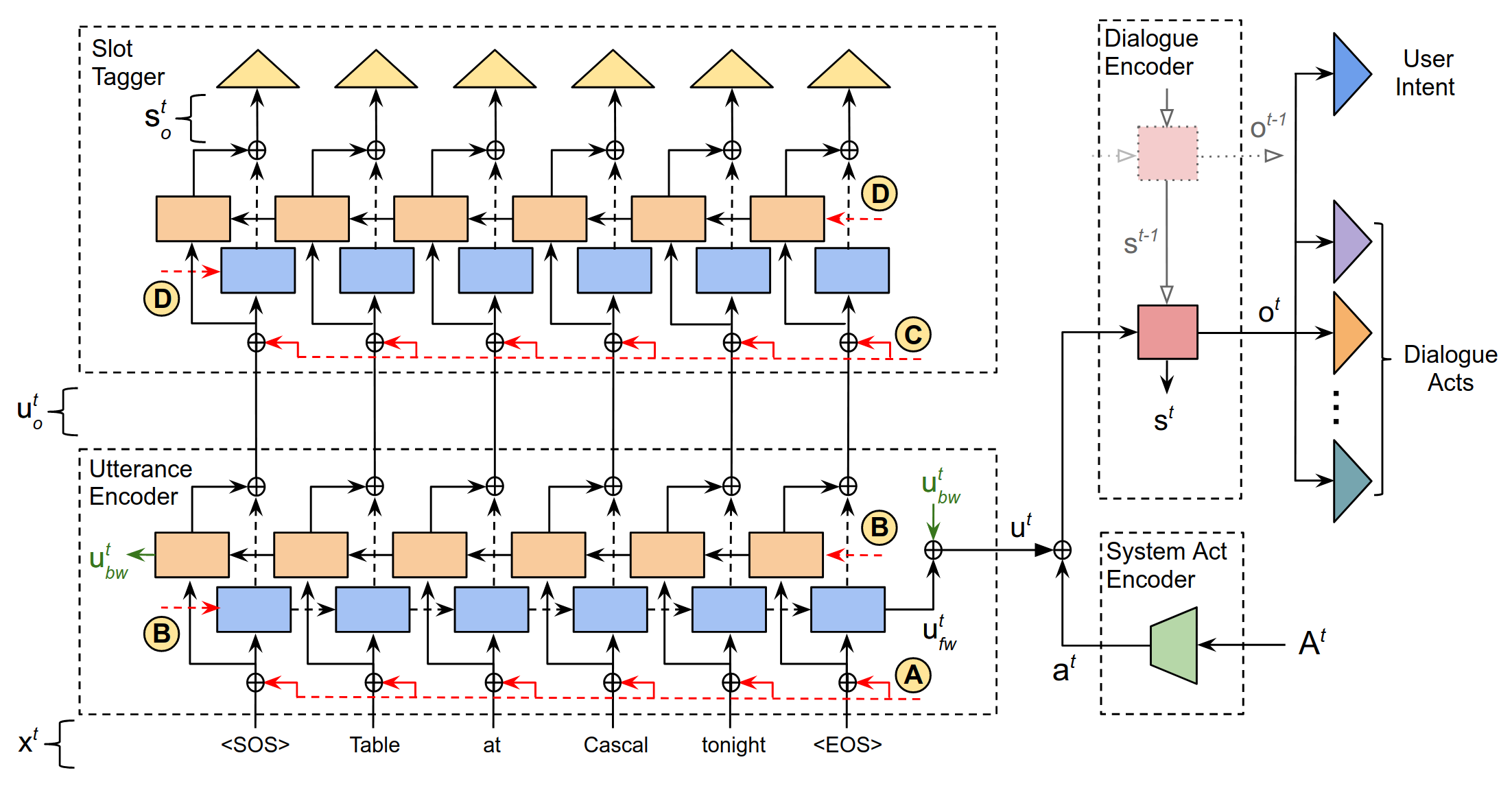}
  \caption{A generalized hierarchical recurrent neural network for joint
  prediction of user intent, dialogue acts (Section \ref{sec:intent}) and slot
  spans (Section \ref{sec:slottag}). Context vectors can be fed as additional
  RNN inputs (positions $A$ and $C$) or can be used to initialize the RNN hidden
  states (positions $B$ and $D$).}
  \label{fig:encoder}
\end{figure*}

Let a dialogue be a sequence of $T$ turns, each turn containing a user utterance
$U^t$  and a set of dialogue acts $A^t$ corresponding to the preceding system
utterance. Figure \ref{fig:encoder} gives an overview of our model architecture.
For a new turn $t$, we use the system act encoder (Section
\ref{sec:act-encoder}) to obtain a vector representation $a^t$ of all system
dialogue acts $A^t$. We also use the utterance encoder (Section
\ref{sec:utterance-encoder}) to generate the user utterance encoding $u^t$ by
processing the user utterance token embeddings $x^t$.

The dialogue encoder (Section \ref{sec:dialog-encoder}) summarizes the content
of the dialogue by using $a^t$, $u^t$, and its previous hidden state $s^{t-1}$
to generate the dialogue context vector $o^t$, and also update its hidden state
$s^t$. The dialogue context vector is then used for user intent classification
and dialogue act classification (Section \ref{sec:intent}). The utterance
encoder also generates updated token embeddings, which are used by the slot
tagger (Section \ref{sec:slottag}) to identify the slot values present in the
user utterance.

Both the utterance encoder and slot tagger use bidirectional RNNs. In addition
to the aforementioned inputs, both RNNs allow for additional inputs (positions
$A$ and $C$ in Figure \ref{fig:encoder}) and external initialization of hidden
states (positions $B$ and $D$ in Figure \ref{fig:encoder}), to incorporate
context in our model. In the following sections, we describe each of these
components in detail.

\subsection{System Act Encoder}
\label{sec:act-encoder}
The system act encoder encodes the set of dialogue acts $A^t$ at turn $t$ into
a vector $a^t$ invariant to the order in which acts appear. This contrasts with
a system utterance-based representation, which imposes an implicit ordering on
the underlying acts.

Each system dialogue act contains an act type and optional slot and value
parameters. We categorize the dialogue acts into two broad types - acts with an
associated slot (and possibly a slot value i.e. \texttt{request(time)},
\texttt{negate(time=`6 pm')}), and acts without (e.g. \texttt{greeting}). Note
that the same dialogue act can appear in the dialogue with or without an
associated slot (\texttt{negate(time=`6 pm')} versus \texttt{negate}).

For each slot type $s$ in our slot vocabulary, we define a binary vector
${a}_{slot}^t(s)$ of size $|A_{sys}|$, where $A_{sys}$ is the set of all system
act types, indicating the presence of each system act type with that slot
associated, ignoring slot values for tractability. Similarly, we define a binary
vector $a^t_{ns}$ of the same size $|A_{sys}|$ indicating the presence of each
system act without any slot associated. For each slot $s$, we also define an
embedding $e_s$. The final encoding $a_t$ is obtained from these vectors after a
shared feedforward layer on the slot-associated act features, followed by
averaging over the set of slots $S^t$ mentioned so far, concatenating with the
no-slot act features and a second feedforward layer, as in equations
\ref{eqn:act-1} - \ref{eqn:act-4}. Parameters $W_1^a$, $W_2^a$, $b_1^a$, $b_2^a$
and slot embeddings $e_s$ are trainable; $\oplus$ denotes concatenation.

\vspace{-10pt}
\begin{gather}
  \label{eqn:act-1}
    {a'}_{slot}^t(s) = a^t_{slot}(s) \oplus e_s \\
  \label{eqn:act-2}
    {a''}_{slot}^t(s) = ReLU(W_1^a \cdot {a'}^t_{slot}(s) + b_1^a) \\
  \label{eqn:act-3} a_{comb}^t = \Big(\frac{1}{|S^t|}\sum_{s\in S^t}
    {a''}_{slot}^t(s)\Big) \oplus a_{ns}^t \\ \label{eqn:act-4} a^t = ReLU(W_2^a
    \cdot a_{comb}^t + b_2^a)
\end{gather}

\subsection{Utterance Encoder}
\label{sec:utterance-encoder}
The user utterance encoder takes in the list of user utterance tokens as input.
Special tokens \texttt{SOS} and \texttt{EOS} are added at the beginning and end
of the token list. Let $x^t = \{x_m^t \in \mathbb{R}^{u_d}, \forall \, 0 \leq
m < M^t\}$ denote the utterance token embeddings, $M^t$ being the number of
tokens in the user utterance for turn $t$. We use a single layer bi-directional
RNN ~\cite{schuster1997bidirectional} using GRU cell \cite{cho2014learning} with
state size $d_u$ to encode the user utterance.

\vspace{-4pt}
\begin{equation}
  \label{eqn:utt-1}
    u^t, u_o^t = {BRNN}_{GRU}(x^t)
\end{equation}

The user utterance encoder outputs embedded representations $u^t \in
\mathbb{R}^{2d_u}$ of the user utterance and $u_o^t = \{u_{o, m}^t \in
\mathbb{R}^{2d_u}, 0 \leq m < M^t\}$ of the individual utterance tokens,
obtained by concatenating the final states and the intermediate outputs of the
forward and backward RNNs respectively.

\subsection{Dialogue Encoder}
\label{sec:dialog-encoder}
The dialogue encoder incrementally generates the embedded representation of the
dialogue context at every turn. We implement the dialogue encoder using a
unidirectional GRU RNN, with each timestep corresponding to a dialogue turn. As
shown in Figure \ref{fig:encoder}, it takes $a^t \oplus u^t$ and its previous
state  $s^{t-1}$ as inputs and outputs the updated state $s^t$ and the encoded
representation of the dialogue context $o^t$ (identical for a GRU RNN). This
method of encoding context is more efficient than other state of the art
approaches like memory networks which process multiple utterances from the
history to process each turn.

\subsection{Intent and Dialogue Act Classification}
\label{sec:intent}
The user intent helps to identify the APIs/databases which the dialogue system
should interact with. Intents are predicted at each turn so that a change of
intent during the dialogue can be detected. We assume that each user utterance
contains a single intent and predict the distribution over all intents at each
turn, $p_i^t$, using equation \ref{eqn:intent-1}. On the other hand, dialogue
act classification is defined as a multi-label binary classification problem to
model the presence of multiple dialogue acts in an utterance. Equation
\ref{eqn:intent-2} is used to calculate $p_a^t$, where $p_a^t(k)$ is the
probability of presence of dialogue act $k$ in turn $t$.

\vspace{-10pt}
\begin{gather}
  \label{eqn:intent-1}
    p_i^t = softmax(W_i \cdot o^t + b_i) \\
  \label{eqn:intent-2}
    p_a^t = sigmoid(W_a \cdot o^t + b_a)
\end{gather}

In the above equations $\dim(p_i^t) = |I|$, $W_i \in \mathbb{R}^{d \times |I|}$,
$W_a \in \mathbb{R}^{d \times |A_u|}$,  $b_i \in \mathbb{R}^{|I|}$, and $b_a \in
\mathbb{R}^{|A_u|}$, $I$ and $A_u$ denoting the user intent and dialogue act
vocabularies respectively and $d = \dim(o^t)$. During inference, we predict
$argmax(p_i^t)$ as the intent label and all dialogue acts with probability
greater than $t_u$ are associated with the utterance, where $0 < t_u < 1.0$ is a
hyperparameter tuned using the validation set.

\subsection{Slot Tagging}
\label{sec:slottag}
Slot tagging is the task of identifying the values for different slots present
in the user utterance. We use the $IOB$ (inside-outside-begin) tagging scheme
(Figure \ref{fig:clu-example}) to assign a label to each token
\cite{tjong2000introduction}. The slot tagger takes the token embeddings output
by the utterance encoder as input and encodes them using a bidirectional RNN
~\cite{schuster1997bidirectional} using LSTM cell \cite{hochreiter1997long} with
hidden state size $d_s$ to generate token embeddings $s^t_o = \{s_{o,m}^t \in
\mathbb{R}^{2d_s}, 0 \leq m < M^t$, $M^t$ being the number of user utterance
tokens in turn $t$. We use an LSTM cell instead of a GRU because it gave better
results on the validation set. For the $m$\textsuperscript{th} token, we use the
token vector $s_{o,m}^t$ to obtain the distribution across all $2|S| + 1$ $IOB$
slot labels using equation \ref{eqn:attention-4}, $|S|$ being the total number
of slot types. During inference, we predict $argmax(p_{s, m}^t)$ as the slot
label for the $m$\textsuperscript{th} token.

\vspace{-10pt}
\begin{gather}
  \label{eqn:attention-4}
    p_{s, m}^t = softmax(W_s \cdot s^t_{o, m} + b_s)
\end{gather}

\section{Experiments}
\label{sec:experiments}

\begin{table*}[h]
  \caption{SLU results on test sets with baselines and our proposed architecture
  variants, when trained on Sim-M + Sim-R. For each dataset, the columns
  indicate the intent accuracy, dialogue act F1 score, slot chunk F1 score and
  frame accuracy, in that order. The Config column indicates the best obtained
  config for feeding context vectors for each experiment.}
  \label{tab:experiments}
  \centering
  \hspace{-1pt}
  \begin{tabularx}{\textwidth}{|p{2.25cm} | p{0.3cm} p{0.3cm} | Y Y Y Y | Y Y Y Y | Y Y Y Y|}\hline
    \textbf{Model} & \multicolumn{2}{c|}{\textbf{Config}} & \multicolumn{4}{c|}{\textbf{Sim-R Results}} & \multicolumn{4}{c|}{\textbf{Sim-M Results}} & \multicolumn{4}{c|}{\textbf{Overall Results}}\\\hline
    & & & Intent & Act & Slot & Frame & Intent & Act & Slot & Frame & Intent & Act & Slot & Frame\\
    & \multicolumn{2}{c|}{$a^t$ \hfill $o^{t-1}$} & Acc & F1 & F1 & Acc & Acc & F1 & F1 & Acc & Acc & F1 & F1 & Acc \\\hline
    1. NoContext & - & - & 83.61 & 87.13 & 94.24 & 65.51 & 88.51 & 93.49 & 86.91 & 62.17 & 84.76 & 89.03 & 92.01 & 64.56\\
    2. PrevTurn & - & - & 99.37 & 90.10 & 94.96 & 86.93 & 99.12 & 93.58 & 88.63 & 77.27 & 99.31 & 91.13 & 93.06 & 84.19\\
    3. MemNet-6 & - & - & 99.75 & 92.90 & 94.42 & 88.33 & 99.12 & 95.71 & 89.76 & 79.11 & 99.68 & 93.74 & 93.03 & 85.71\\
    4. MemNet-20 & - & - & 99.67 & 95.67 & 94.28 & \textbf{89.52} & 98.76 & 96.25 & 90.70 & 80.35 & 99.29 & 95.85 & 93.21 & 86.92\\
    5. SDEN-6 & - & - & 99.76 & 93.14 & 95.83 & 88.74 & 99.74 & 95.02 & 88.60 & 79.11 & 99.76 & 93.70 & 93.66 & 86.01\\
    6. SDEN-20 & - & - & 99.84 & 94.43 & 94.81 & 89.46 & 99.60 & 97.56 & 90.93 & \textbf{82.55} & 99.81 & 95.38 & 93.65 & \textbf{87.50}\\\hline
    7. $a^t$ only, No DE & B & - & 99.62 & 93.21 & 95.53 & 87.63 & 99.12 & 96.00 & 87.30 & 75.44 & 99.48 & 94.04 & 93.07 & 84.17\\
    8. $a^t$ only & D & - & 99.98 & 95.42 & 95.38 & \textbf{89.26} & 99.71 & 96.35 & 91.58 & 83.36 & 99.92 & 95.70 & 94.22 & \textbf{87.58}\\
    9. $o^{t-1}$ only & - & D & 99.83 & 94.44 & 94.18 & 87.63 & 99.27 & 96.66 & 91.88 & 86.80 & 99.67 & 95.11 & 93.46 & 87.40\\
    10. $a^t$ and $o^{t-1}$ & C & D & 99.65 & 92.71 & 94.70 & 87.54 & 99.27 & 96.11 & 93.73 & \textbf{86.88} & 99.54 & 93.74 & 94.40 & 87.35\\\hline
  \end{tabularx}
\end{table*}

We use two representations of dialogue context: the dialogue encoding vector
$o^{t-1}$ encodes all turns prior to the current turn and the system intent
vector $a^t$ encodes the current turn system utterance. Thus, $o^{t-1}$ and
$a^t$ together encode the entire conversation observed till the user utterance.
These vectors can be fed as inputs at multiple places in the SLU model. In this
work, we identify four positions to feed context i.e. positions A through D in
Figure \ref{fig:encoder}. Positions A and C feed context vectors as additional
inputs at each RNN step whereas positions B and D use the context vectors to
initialize the hidden state of the two RNNs after a linear projection to the
hidden state dimension. We experiment with the following configurations for
integrating dialogue context in our framework:

\begin{enumerate}[leftmargin=10pt, topsep=0pt, noitemsep]
  \item \textit{\textit{$a^t$ only, No DE:}} We feed $a^t$, the system act
  encoding, in one of positions A-D, omit the dialogue encoder, and instead
  use $u^t$, the utterance encoder's final state, for intent and act prediction.
  The best model for this configuration, as evaluated on the validation set, had
  $a^t$ fed in position B, and test set results for this model are reported in
  row 7 of Table \ref{tab:experiments}.

  \item \textit{\textit{$a^t$ only:}} We feed $a^t$ into the dialogue encoder,
  and to one of the positions A-D. Row 8 of Table \ref{tab:experiments} contains
  results for the best model for this configuration, which had $a^t$ fed in
  position D of the slot tagger.

  \item \textit{\textit{$o^{t-1}$ only:}} We feed $a^t$ into the dialogue
  encoder and $o^{t-1}$, the dialogue encoding from the previous turn, into the
  slot tagger at positions C or D. Row 9 of Table \ref{tab:experiments} shows
  results for the best model with $o^{t-1}$ fed in position D.

  \item \textit{\textit{$a^t$ and $o^{t-1}$:}} We feed $a^t$ into the dialogue
  encoder, and $o^{t-1}$ and $a^t$ independently into one of positions C and D,
  4 combinations in total. Row 10 of Table \ref{tab:experiments} shows results
  for the best model with $a^t$ fed in position C and $o^{t-1}$ in position D.
\end{enumerate}

\subsection{Dataset}
\label{sec:dataset}
We obtain dialogues from the Simulated Dialogues dataset \footnote{The dataset
is available at http://github.com/google-research-datasets/simulated-dialogue},
described in ~\cite{shah2017building}. The dataset has dialogues from restaurant
(Sim-R, 11234 turns in 1116 training dialogues) and movie (Sim-M, 3562 turns in
384 training dialogues) domains and a total of three intents. The dialogues in
the dataset consist of 12 slot types and 21 user dialogue act types, with 2 slot
types and 12 dialogue acts shared between Sim-R and Sim-M. One challenging
aspect of this dataset is the prevalence of unseen entities. For instance, only
13\% of the movie names in the validation and test sets are also present in the
training set.

\subsection{Baselines}
We compare our models' performance with the following four baseline models:

\begin{enumerate}[leftmargin=10pt, topsep=0pt, noitemsep]
  \item \textit{NoContext:} A two-layer stacked bidirectional RNN using GRU and
  LSTM cells respectively, and no context.

  \item \textit{PrevTurn:} This is similar to the \textit{NoContext} model.
  with a different bidirectional GRU layer encoding the previous system turn,
  and this encoding being input to the slot tagging layer of encoder i.e.
  position C in Figure \ref{fig:encoder}.

  \item \textit{MemNet:} This is the system from \cite{chen2016end2end}, using
  cosine attention. For this model, we report metrics with models trained with
  memory sizes of 6 and 20 turns. A memory size of 20, while making the model
  slower, enables it to use the entire dialogue history for most of the
  dialogues.

  \item \textit{SDEN:} This is the system from \cite{bapna2017sequential} which
  uses a bidirectional GRU RNN for combining memory embeddings. We report
  metrics for models with memory sizes 6 and 20.
\end{enumerate}

\subsection{Training and Evaluation}
\label{sec:training}
We use sigmoid cross entropy loss for dialogue act classification (since it is
modeled as a multilabel binary classification problem) and softmax cross entropy
loss for intent classification and slot tagging. During training, we minimize
the sum of the three constituent losses using the ADAM optimizer
\cite{kingma2014adam} for 150k training steps with a batch size of 10 dialogues.

To improve model performance in the presence of out of vocabulary (OOV) tokens
arising from entities not present in the training set, we randomly replace
tokens corresponding to slot values in user utterance with a special OOV token
with a value dropout probability that linearly increases during training.

To find the best hyperparameter values, we perform grid search over the token
embedding size ($\in \{64,128,256\}$), learning rate ($\in [0.0001, 0.01]$),
maximum value dropout probability ($\in [0.2, 0.5]$) and the intent prediction
threshold ($\in \{0.3, 0.4, 0.5\}$), for each model configuration listed in
Section \ref{sec:experiments}. The utterance encoder and slot tagger layer sizes
are set equal to the token embedding dimension, and that of the dialogue encoder
to half this dimension. In Table \ref{tab:experiments}, we report intent
accuracy, dialogue act F1 score,
slot chunk F1 score \cite{tjong2000introduction} and frame accuracy on the test
set for the best runs for each configuration in Section \ref{sec:experiments}
based on frame accuracy on the combined validation set, to avoid overfitting. A
frame is considered correct if its predicted intent, slots and acts are all
correct.

\section{Results and Discussion}
\label{sec:results}
Table \ref{tab:experiments} compares the baseline models with different variants
of our model. We observe that our models compare favorably to the state of the
art MemNet and SDEN baselines. The use of context plays a crucial role across
all datasets and tasks, especially for intent and dialogue act classification,
giving an improvement of $\sim$15\% and $\sim$5\% respectively across all
configurations. For all subsequent discussion, we concentrate on frame accuracy
since it summarizes the performance across all tasks.

An important consideration is the computational efficiency of the compared
appraoches: memory network-based models are expensive to run, since they process
multiple utterances from the dialogue history at every turn. In contrast, our
approach only adds a two-layer feedforward network (the system act encoder) and
one step of a GRU cell (for the dialogue encoder) per turn to encode all
context. Empirically, MemNet-6 and MemNet-20 experiments took roughly 4x and 12x
more time to train respectively than our slowest model containing both the
system act encoder and the dialogue encoder, on our training setup. SDEN runs
are slower than their MemNet counterparts since they use RNNs for combining
memory embeddings. In addition to being fast, our models generalize better on
the smaller Sim-M dataset, suggesting that memory network-based models tend to
be more data intensive.

Two interesting experiments to compare are rows 2 and 7 i.e. ``PrevTurn" and
``$a^t$ only, No DE"; they both use context only from the previous system
utterance/acts, discarding the remaining turns. Our system act encoder,
comprising only a two-layer feedforward network, is in principle faster than the
bidirectional GRU that ``PrevTurn" uses to encode the system utterance. This
notwithstanding, the similar performance of both models suggests that using
system dialogue acts for context is a good alternative to using the
corresponding system utterance.

Table \ref{tab:experiments} also lists the best configurations for feeding
context vectors $a^t$ and $o^{t-1}$. In general, we observe that feeding context
vectors as initial states to bidirectional RNNs (i.e. position D for slot
tagging plus a side input to the dialogue encoder, or position B for all tasks
in case of no dialogue encoder) yields better results than feeding them as
additional inputs at each RNN step (positions C and A). This may be caused by
the fact that our context vectors do not vary with the user utterance tokens,
because of which introducing them repeatedly is likely redundant. For each
experiment in Section \ref{sec:experiments}, the differences between the varying
context position combinations are statistically significant, as determined by
McNemar's test with $p < 0.05$.

Another interesting observation is that using $o^{t-1}$ as compared to $a^t$ as
additional context for the slot tagger does not improve slot tagging
performance. This indicates a strong correspondence between the slots present in
the system acts and those mentioned in the user utterance i.e. the user is often
responding directly to the immediately prior system prompt, thereby reducing the
dependence on context from the previous turns for the slot tagging task.

To conclude, we present a fast and efficient approach to encoding context for
SLU. Avoiding the significant per-turn overhead of memory networks, our method
accumulates dialogue context one turn at a time, resulting in a faster and more
generalizable model without any loss in accuracy. We also demonstrate that using
system dialogue acts is an efficientalternative to using system utterances for
context.

\bibliographystyle{IEEEtran}
\bibliography{bibliography}
\end{document}